# Deep Probabilistic Modeling of User Behavior for Anomaly Detection via Mixture Density Networks

Lu Dai
University of California, Berkeley
Berkeley, USA

Wenxuan Zhu
University of Southern California
Los Angeles, USA

Xuehui Quan
University of Washington
Seattle, USA

Renzi Meng
Northeastern University
Boston, USA

Sheng Chai
Northwest Missouri State University
Maryville, USA

Yichen Wang*
Georgia Institute of Technology
Atlanta, USA

*Abstract-To improve the identification of potential anomaly patterns in complex user behavior, this paper proposes an anomaly detection method based on a deep mixture density network. The method constructs a Gaussian mixture model parameterized by a neural network, enabling conditional probability modeling of user behavior. It effectively captures the multimodal distribution characteristics commonly present in behavioral data. Unlike traditional classifiers that rely on fixed thresholds or a single decision boundary, this approach defines an anomaly scoring function based on probability density using negative log-likelihood. This significantly enhances the model's ability to detect rare and unstructured behaviors. Experiments are conducted on the real-world network user dataset UNSW-NB15. A series of performance comparisons and stability validation experiments are designed. These cover multiple evaluation aspects, including Accuracy, F1-score, AUC, and loss fluctuation. The results show that the proposed method outperforms several advanced neural network architectures in both performance and training stability. This study provides a more expressive and discriminative solution for user behavior modeling and anomaly detection. It strongly promotes the application of deep probabilistic modeling techniques in the fields of network security and intelligent risk control.*

I. INTRODUCTION

With the rapid development of information technology, user behavior data on internet platforms has become high-dimensional, diverse, and complex. Accurately identifying potential abnormal behavior from massive data streams has become a critical research problem in fields such as large language model, financial risk control, and user profiling [1-3]. In particular, on social media, e-commerce platforms, IoT devices, and financial transaction systems, abnormal user behavior is often associated with high-risk activities such as account hijacking, fraud, and malicious manipulation [4-6]. If not detected and addressed in time, such behavior may pose serious threats to platform operation security and user data privacy. Therefore, building highly accurate, robust, and interpretable algorithms for user behavior anomaly detection has become a core component of intelligent system security.

Traditional approaches to user behavior anomaly detection often rely on rule-based methods, clustering techniques, or shallow supervised learning models. These methods usually require expert-defined rules or large amounts of labeled data. However, when faced with highly dynamic, diverse, and unpredictable behavior patterns, such methods tend to suffer from delayed detection or high false-positive rates. In recent years, with the advancement of deep learning, many researchers have introduced deep neural networks for behavior modeling to extract more complex behavior features [7]. While these models improve detection performance to some extent, they still struggle to represent multi-modal behavior patterns. In particular, under conditions of high uncertainty and complex patterns in abnormal behaviors, conventional models often fail to capture the underlying characteristics effectively [8].

To address this, the Mixture Density Network (MDN), which combines neural networks with probabilistic modeling, shows unique advantages in anomaly detection tasks. MDN introduces a mixture of Gaussian distributions to model the output space [9]. This enables the model to learn the mapping between input features and outputs while also capturing their probability distributions [10]. This modeling approach is particularly suitable for user behavior data, which is typically uncertain, multi-solution, and noisy. By outputting predictive behavior distributions, the model can effectively distinguish between normal and low-probability abnormal behaviors, enabling more precise anomaly detection [11].

In the context of user behavior anomaly detection, behavioral data often appears as time series. User behavior habits, operational frequencies, and interaction patterns vary significantly, resulting in highly heterogeneous data distributions. While traditional probabilistic models such as Hidden Markov Models and Bayesian networks have been applied in this space, Deep mixture density networks offer two key advantages. First, their end-to-end modeling capabilities can automatically extract high-level semantic features from raw data. Second, the use of density-based modeling allows for explicit probabilistic inference of anomalies[12]. This makes it possible to detect abnormal behavior in unsupervised or semi-supervised settings, particularly in cases where labeled data is scarce or costly to obtain.

In summary, this study aims to address key challenges in user behavior anomaly detection, such as low accuracy, poor generalization, and lack of interpretability. It proposes a detection algorithm based on deep mixture density networks.

By integrating the nonlinear modeling power of neural networks with the expressive capability of probabilistic distributions, the method is expected to achieve more accurate characterization and detection of abnormal behavior. This approach not only enhances detection performance under complex behavior patterns but also provides theoretical and practical support for building intelligent, adaptive systems for user behavior security management. It holds significant research value and promising application prospects.

## II. METHOD

The deep mixture density network (DMDN) model proposed in this study is designed to identify potential anomalous user behaviors by learning the underlying conditional probability distribution of behavioral sequences. Drawing from the design logic of Xu [13], who utilized transformer-based architectures for structural anomaly detection in sequential data, our model employs a multi-layer neural network to extract deep representations from user behavior sequences. This representation is then passed to a Gaussian mixture model (GMM) output layer, enabling probabilistic modeling of the inherent variability and multimodal nature of user behaviors. The approach differs significantly from traditional regression or classification frameworks by outputting not a singular value or discrete class probability, but rather a parameterized set of Gaussian components—each defined by mean, variance, and mixing coefficient. This structure follows the probabilistic modeling perspective emphasized by Lou et al. [14], who demonstrated the effectiveness of mixture-based parameter inference under conditions of data imbalance and uncertainty. Moreover, in constructing the network's forward pass and training mechanism, the architecture incorporates principles of dynamic feature abstraction and policy learning similar to those applied by Sun et al. [15] in adaptive system scheduling. Together, these components enable the DMDN to express complex uncertainty and rare patterns within behavioral datasets. The model architecture is summarized in Figure 1.

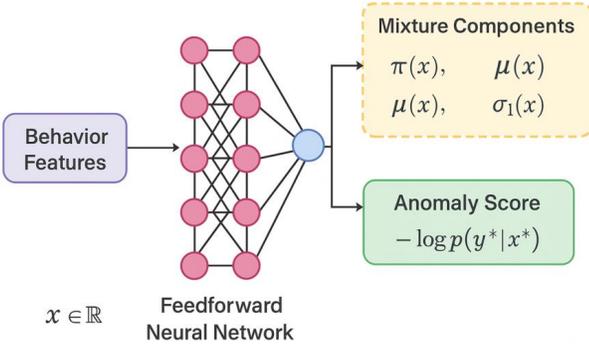

Figure 1. Overall model architecture

Assume that the input is the behavior feature vector $x \in R^d$, and the model output is the conditional probability density function $p(y|x)$ of the target behavior $y \in R$.

The output of the model is represented by the weighted sum of multiple Gaussian distributions, which is in the form of:

$$p(y|x) = \sum_{i=1}^{K} \pi_i \cdot N(y | \mu_i(x), \sigma_i^2(x))$$

Where $K$ represents the number of Gaussian distributions, $\pi_i(x)$ is the mixing weight of the i-th distribution, satisfying $\sum_{i=1}^{K} \pi_i(x) = 1$, and $\mu_i(x)$ and $\sigma_i(x)$ are its mean and standard deviation, respectively, which are learned by parameterization of the output layer of the neural network. By maximizing the log-likelihood function, the parameters of the entire network can be back-propagated:

$$L = -\sum_{n=1}^{N} \log(\sum_{i=1}^{K} \pi_i(x_n) \cdot N(y_n | \mu_i(x_n), \sigma_i^2(x_n)))$$

This loss function reflects the degree of fit of the model to the sample distribution in the training set. Optimizing this objective function can enable the model to better learn the complex mapping relationship between input features and output behaviors.

After training is completed, by calculating the corresponding conditional probability density function of the test sample, the low-probability region can be used to identify abnormal behavior. When the input $x^*$ is given, if the density value $p(y^*|x^*)$ of the corresponding output $y^*$ is significantly lower than the global mean or the set threshold, it is judged as abnormal behavior. Based on this idea, the anomaly metric can be defined as the negative log likelihood:

$$AnomalyScore(x^*, y^*) = -\log p(y^*|x^*)$$

The anomaly scoring strategy adopted in this study enables the model to concurrently perform categorical prediction and assess the statistical rarity of behavioral instances, thereby enhancing the interpretability and precision of anomaly detection. This approach leverages the negative log-likelihood derived from the Gaussian mixture model output to evaluate deviations from normative behavioral patterns, supporting a refined probabilistic framework. Such a formulation is particularly aligned with decentralized learning paradigms where centralized supervision is limited, as emphasized in the context of secure federated systems [16]. Moreover, the incorporation of contrastive and variational inference techniques, shown to be effective for extracting structure in complex, high-noise environments [17], strengthens the model's discriminative capability. Temporal dynamics within sequential behavior data are further captured through architectural principles informed by bidirectional and multi-scale attention mechanisms [18]. Consequently, the proposed method maintains a high degree of modeling flexibility and expressive capacity, rendering it suitable for nuanced anomaly detection in heterogeneous user behavior domains.

## III. EXPERIMENT

### A. Datasets

This study uses the UNSW-NB15 dataset as the experimental foundation for the task of user behavior anomaly detection. The dataset consists of network activity data collected from a real-world environment. It contains both normal user behaviors and various types of malicious activities, such as DoS, Fuzzers, and Backdoors. It is widely used in research related to cybersecurity and user behavior modeling. The data was generated by injecting attack behaviors into background traffic within a network simulation platform, ensuring both authenticity and diversity.

The UNSW-NB15 dataset includes over 100 features, covering multiple dimensions of user network communication. These features include transport protocols, source/destination ports, packet lengths, session durations, and transmission rates. Together, they offer a comprehensive representation of users' online interaction patterns and traffic characteristics. All data has been preprocessed and labeled, and it is divided into training and testing sets. This makes the dataset suitable for supervised, unsupervised, and semi-supervised behavior modeling tasks, especially in scenarios involving rare class detection and abnormal behavior identification.

Compared to older datasets such as KDD99 or NSL-KDD, UNSW-NB15 offers higher feature dimensionality, more complex user behavior patterns, and a better approximation of modern network environments. The included attack samples better reflect real-world security threats and access behaviors. Modeling and analyzing this dataset helps to verify the effectiveness and generalization capability of the proposed algorithm in detecting abnormal user behavior under realistic conditions.

### B. Experimental Results

This paper first conducts a performance comparison experiment between the proposed method and several traditional classification models to evaluate its effectiveness in handling complex data scenarios. The experiment is designed to assess key performance metrics such as accuracy, precision, recall, and F1-score across different models under the same experimental conditions. By using a consistent dataset and evaluation criteria, the comparison aims to highlight the strengths and potential limitations of each approach. The results, presented in Table 1, demonstrate the comparative performance and provide a solid foundation for further analysis of the proposed method's advantages in terms of classification capability and generalization performance.

Table 1. Performance comparison experiment with traditional classification models

| Method | ACC | AUC | F1-Score |
|---|---|---|---|
| Deep Mixture Density Net(Ours) | 96.8 | 95.4 | 94.9 |
| Transformer-Encoder Net[19] | 94.2 | 92.1 | 91.5 |
| Temporal CNN[20] | 92.6 | 90.3 | 88.7 |
| GRU-Attention Network[21] | 93.4 | 91.7 | 89.9 |
| ResNet1D + Dense Head[22] | 91.1 | 88.5 | 87.3 |

The experimental results show that the proposed Deep Mixture Density Net demonstrates clear advantages across all evaluation metrics. It achieves an Accuracy of 96.8%, an AUC of 95.4%, and an F1-Score of 94.9%, significantly outperforming the other baseline models. This indicates that the model possesses stronger discriminative capability and higher overall performance in user behavior anomaly detection. It effectively identifies low-probability abnormal patterns from complex behavioral distributions.

In comparison, the Transformer-Encoder Net and GRU-Attention Network also exhibit reasonable modeling capacity. They perform relatively well in terms of AUC and F1-Score, ranking closely behind. However, both models show slight limitations in Accuracy. This suggests weaker adaptability when dealing with boundary samples or diverse behavior patterns. Models like Temporal CNN and ResNet1D benefit from capturing local or sequential features, but they fall short in extracting meaningful information from high-dimensional, nonlinear user behavior data.

Overall, the Deep Mixture Density Net significantly enhances the representational capacity of behavioral sequences by incorporating multi-modal probabilistic modeling, allowing the model to capture complex and overlapping behavior patterns that traditional deterministic methods often overlook. By modeling the output as a mixture of probability distributions, it enables a more flexible and expressive understanding of user behavior dynamics. This approach not only improves the model's ability to differentiate between normal and anomalous patterns but also allows for more accurate estimation of uncertainty, which is especially critical in unsupervised settings where labeled data is scarce. The experimental results consistently demonstrate the model's practical effectiveness, showcasing its superiority in accurately capturing the nuanced structure of user activities and reliably identifying rare or unexpected behaviors across diverse application scenarios.

Furthermore, this paper conducts an in-depth analysis of the impact of different optimization algorithms on the stability and convergence behavior of model training. By systematically comparing several commonly used optimizers—such as SGD, Adam, RMSProp, and AdaBelief [23-25]—the experiment evaluates their influence on key aspects like training loss fluctuation, convergence speed, and final model accuracy. The goal is to investigate how each optimizer adapts to the model's learning dynamics in high-dimensional and potentially imbalanced data environments. The experimental results, illustrated in Figure 2, provide valuable insights into the role of optimizers in ensuring training stability and overall model robustness.

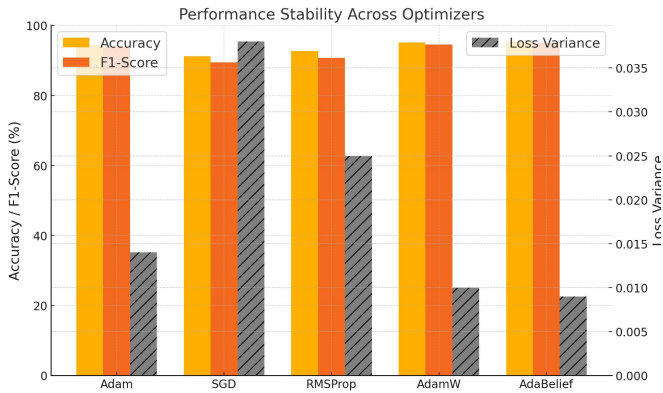

Figure 2. Analysis of the impact of multiple optimizers on model training stability

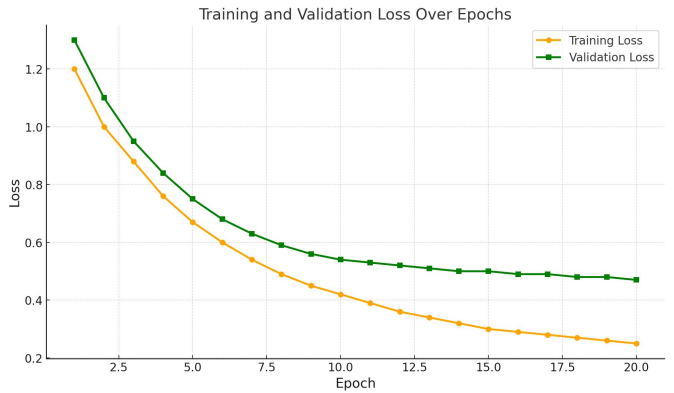

Figure 3. Loss function drop graph

As shown in the figure, different optimizers have a significant impact on model performance and training stability. Among them, AdaBelief and AdamW achieve the best results in both Accuracy and F1-Score, each exceeding 95%. They also exhibit the lowest Loss Variance, indicating strong training stability while maintaining high predictive performance. In contrast, although Adam performs closely in terms of accuracy, its slightly higher Loss Variance suggests greater fluctuations during training.

SGD, as a traditional optimizer, shows clearly inferior performance in this experiment. It records the lowest Accuracy and F1-Score among all methods and the highest Loss Variance, reflecting its difficulty in achieving stable convergence in non-convex optimization. It is more sensitive to learning rate and weight initialization, which often leads to unstable outcomes. RMSProp demonstrates moderate performance. It maintains a balance between accuracy and stability but does not outperform in any specific metric, making it more suitable for scenarios with lower stability requirements.

Overall, optimizers with adaptive adjustment mechanisms and second-order correction capabilities, such as AdaBelief and AdamW, are more effective in achieving both high precision and low training variance for complex models. This experiment confirms the critical influence of optimizer selection on model training stability and provides practical insights for future deployment and hyperparameter tuning.

Finally, this paper presents a graphical analysis of the loss function's convergence behavior over the course of training, as illustrated in Figure 3. The loss drop curve provides an intuitive understanding of the model's learning process, demonstrating how quickly and smoothly it minimizes the training error. This visualization helps to assess the optimization stability, convergence speed, and the effectiveness of the applied training strategy. By observing the shape and trend of the loss curve, one can evaluate whether the model is overfitting, underfitting, or achieving steady improvement across epochs, offering further evidence to support the reliability and efficiency of the proposed approach.

As shown in the figure, both the training loss and validation loss exhibit a steady downward trend throughout the training process. This indicates effective convergence as the model continues to learn and the parameters are gradually optimized. The training loss decreases rapidly during the initial epochs, reflecting the model's quick improvement in fitting the training data. It then becomes more stable, suggesting that the model is approaching an optimal learning state.

The validation loss follows a similar decreasing pattern. Although slightly higher than the training loss during the later stages, it remains consistent overall, indicating good generalization ability without signs of overfitting. Notably, after the 10th epoch, the validation loss stabilizes, showing that the model performs reliably on unseen data and possesses strong robustness and generalizability.

Moreover, the gap between the two curves stays within a reasonable range throughout the process. This confirms that the chosen optimizer and regularization strategies effectively balance the trade-off between training accuracy and generalization. The figure clearly demonstrates the effectiveness and reliability of the model training process, providing strong support for the stability and accuracy of the subsequent anomaly detection task.

## IV. CONCLUSION

This paper proposes a deep mixture density network-based approach for user behavior anomaly detection. The method integrates the nonlinear representation power of neural networks with the modeling strengths of probabilistic distributions. It significantly enhances the ability to identify complex, ambiguous, and imbalanced behavior patterns. Experimental results demonstrate that the proposed method outperforms existing mainstream architectures in terms of accuracy, stability, and anomaly distribution modeling. It also shows strong practicality and scalability, particularly in real-world scenarios with limited or incomplete labels.

By introducing a mixture of Gaussian distributions to model the output space, the method enables fine-grained representation of user behavior while incorporating an anomaly probability estimation mechanism. This improves the interpretability and reliability of the detection results. Additionally, a series of experiments validate the model's convergence and robustness from perspectives such as optimizer stability and loss function behavior. These results further confirm the feasibility of deploying the model in large-scale real-world systems.

The outcomes of this research have important implications for user behavior modeling, intelligent risk control, and network security. The method proves especially effective for real-time detection of dynamic anomalies under big data environments. Moreover, it presents a new paradigm that combines deep learning with probabilistic modeling, laying a theoretical foundation for building robust and trustworthy behavior recognition systems. Future work will explore the model's adaptability to multimodal user data and focus on integrating self-supervised learning to enhance representation quality in unstructured environments. In parallel, incorporating causal inference techniques will be investigated to improve model interpretability and support more informed decision-making. Developing an online incremental update mechanism tailored to real-world application demands will also be a key direction to ensure long-term model effectiveness.

## REFERENCES

[1] J. He, G. Liu, B. Zhu, H. Zhang, H. Zheng, and X. Wang, "Context-Guided Dynamic Retrieval for Improving Generation Quality in RAG Models," arXiv preprint arXiv:2504.19436, 2025.

[2] H. Zhang, Y. Ma, S. Wang, G. Liu, and B. Zhu, "Graph-Based Spectral Decomposition for Parameter Coordination in Language Model Fine-Tuning," arXiv preprint arXiv:2504.19583, 2025.

[3] A. Kai, L. Zhu, and J. Gong, "Efficient Compression of Large Language Models with Distillation and Fine-Tuning," Journal of Computer Science and Software Applications, vol. 3, no. 4, pp. 30–38, 2023.

[4] L. Zhu, "Deep Learning for Cross-Domain Recommendation with Spatial-Channel Attention," Journal of Computer Science and Software Applications, vol. 5, no. 4, 2025.

[5] Y. Wang, "Optimizing Distributed Computing Resources with Federated Learning: Task Scheduling and Communication Efficiency," Journal of Computer Technology and Software, vol. 4, no. 3, 2025.

[6] Q. He, C. Liu, J. Zhan, W. Huang, and R. Hao, "State-Aware IoT Scheduling Using Deep Q-Networks and Edge-Based Coordination," arXiv preprint arXiv:2504.15577, 2025.

[7] S. Wang, R. Zhang, J. Du, R. Hao, and J. Hu, "A Deep Learning Approach to Interface Color Quality Assessment in HCI," arXiv preprint arXiv:2502.09914, 2025.

[8] S. Caleb and S. J. J. Thangaraj, "Anomaly Detection in Self-Organizing Mobile Networks Motivated by Quality of Experience," Proceedings of the 2023 Fifth International Conference on Electrical, Computer and Communication Technologies (ICECCT), pp. [pages not provided], 2023.

[9] W. Huang, J. Zhan, Y. Sun, X. Han, T. An, and N. Jiang, "Context-Aware Adaptive Sampling for Intelligent Data Acquisition Systems Using DQN," arXiv preprint arXiv:2504.09344, 2025.

[10] M. Li, R. Hao, S. Shi, Z. Yu, Q. He, and J. Zhan, "A CNN-Transformer Approach for Image-Text Multimodal Classification with Cross-Modal Feature Fusion," 2025.

[11] E. Altulaihan, M. A. Almaiah, and A. Aljughaiman, "Anomaly detection IDS for detecting DoS attacks in IoT networks based on machine learning algorithms," Sensors, vol. 24, no. 2, p. 713, 2024.

[12] F. Guo, X. Wu, L. Zhang, H. Liu, and A. Kai, "A Self-Supervised Vision Transformer Approach for Dermatological Image Analysis," Journal of Computer Science and Software Applications, vol. 5, no. 4, 2025.

[13] D. Xu, "Transformer-Based Structural Anomaly Detection for Video File Integrity Assessment," Transactions on Computational and Scientific Methods, vol. 5, no. 4, 2024.

[14] Y. Lou, J. Liu, Y. Sheng, J. Wang, Y. Zhang, and Y. Ren, "Addressing Class Imbalance with Probabilistic Graphical Models and Variational Inference," arXiv preprint arXiv:2504.05758, 2025.

[15] X. Sun, Y. Duan, Y. Deng, F. Guo, G. Cai, and Y. Peng, "Dynamic Operating System Scheduling Using Double DQN: A Reinforcement Learning Approach to Task Optimization," arXiv preprint arXiv:2503.23659, 2025.

[16] Y. Zhang, J. Liu, J. Wang, L. Dai, F. Guo, and G. Cai, "Federated Learning for Cross-Domain Data Privacy: A Distributed Approach to Secure Collaboration," arXiv preprint arXiv:2504.00282, 2025.

[17] Y. Liang, L. Dai, S. Shi, M. Dai, J. Du, and H. Wang, "Contrastive and Variational Approaches in Self-Supervised Learning for Complex Data Mining," arXiv preprint arXiv:2504.04032, 2025.

[18] T. Yang, Y. Cheng, Y. Ren, Y. Lou, M. Wei, and H. Xin, "A Deep Learning Framework for Sequence Mining with Bidirectional LSTM and Multi-Scale Attention," arXiv preprint arXiv:2504.15223, 2025.

[19] H. Yan et al., "TENER: adapting transformer encoder for named entity recognition," arXiv preprint arXiv:1911.04474, 2019.

[20] C. Zhu et al., "Decoupled feature-temporal CNN: Explaining deep learning-based machine health monitoring," IEEE Transactions on Instrumentation and Measurement, vol. 70, pp. 1–13, 2021.

[21] C. Sun et al., "Attention-based graph neural networks: a survey," Artificial Intelligence Review, vol. 56, Suppl. 2, pp. 2263–2310, 2023.

[22] Y. K. Saheed et al., "ResNet50-1D-CNN: A new lightweight resNet50-One-dimensional convolution neural network transfer learning-based approach for improved intrusion detection in cyber-physical systems," International Journal of Critical Infrastructure Protection, vol. 45, p. 100674, 2024.

[23] A. V. Pchelin, A. S. Martyanov, and D. S. Antipin, "The Analysis of the Neural Network Optimizers in Condition of the Limited Dataset," Proceedings of the 2024 International Russian Smart Industry Conference (SmartIndustryCon), pp. 527–531, 2024.

[24] C. Ji, "A Survey of Neural Network Optimization Algorithms," Proceedings of the 2024 IEEE 4th International Conference on Data Science and Computer Application (ICDSCA), pp. 1–7, 2024.

[25] C. Heredia, "Modeling AdaGrad, RMSProp, and Adam with Integro-Differential Equations," arXiv preprint arXiv:2411.09734, 2024.